\documentclass[10pt,conference]{IEEEtran}
\usepackage{graphicx}
\usepackage{booktabs}
\usepackage{multirow}
\usepackage{amsmath,amssymb}
\usepackage{array}
\usepackage{microtype}
\usepackage{xcolor}
\usepackage{tikz}
\usetikzlibrary{arrows.meta,positioning,fit}
\usepackage{url}
\usepackage{cite}

\title{An AI-Based Multi-Stage Approach for Androgenetic Alopecia Assessment from Low-Magnification Scalp Images}
\author{
    \IEEEauthorblockN{Mahmoud Raslan\IEEEauthorrefmark{1},
    Nada Omar\IEEEauthorrefmark{1},
    Omar Khaled\IEEEauthorrefmark{1},
    Tarek Waleed\IEEEauthorrefmark{1},
    Mohamed Hazem\IEEEauthorrefmark{1},
    Rania Mounir\IEEEauthorrefmark{2},}
    \IEEEauthorblockN{Solwan Elsamanoudy\IEEEauthorrefmark{2},
    Ahmed Mourad\IEEEauthorrefmark{2},
    Noura Adel\IEEEauthorrefmark{2},
    Muhammad Rushdi\IEEEauthorrefmark{1}}
    \vspace{0.15cm}
    \IEEEauthorblockA{\IEEEauthorrefmark{1}Faculty of Engineering, Cairo University, Giza, Egypt}
    \IEEEauthorblockA{\IEEEauthorrefmark{2}Dermatology Department, Faculty of Medicine, Cairo University, Cairo, Egypt}
}

\begin{document}
\maketitle

\begin{abstract}
Androgenetic alopecia (AGA) is characterized by patterned follicular miniaturization, increased single-hair follicular units, and altered hair-shaft diameter. We present an automated quantitative scalp-analysis and clinical decision-support framework combining FU localization, ordinal visible-shaft counting, calibrated shaft-width estimation, regional aggregation, and an interpretable rule layer. The clinical cohort comprised 243 patients (127 AGA, 116 non-AGA), while the computer-vision experiments used 160 expert-annotated patients, 2,400 trichoscopic images, and approximately 158,000 FU annotations. Under patient-disjoint evaluation, YOLOv8m achieved test mAP@0.5=0.920 and recall=0.860; EfficientNet-B5 with a support-map channel achieved 87.0\% expert-box count accuracy (macro F1=0.85). A separate 500-image set was processed end-to-end with detector-generated boxes, yielding MAE of 6.56 for follicle detection and 16.59 for follicle classification relative to human-expert annotations. The system is intended to assist, rather than replace, dermatologist interpretation.
\end{abstract}


\section{Introduction}
Androgenetic alopecia (AGA), commonly known as pattern hair loss, is a widespread dermatological condition affecting both males and females worldwide~\cite{ref1,ref2}. The disease is characterized by progressive hair thinning, reduced follicular density, and hair-shaft miniaturization, which may significantly affect patients psychologically and socially. Reliable severity assessment is important for treatment planning and longitudinal monitoring of therapeutic response.

Trichoscopy has become an important non-invasive imaging technique for evaluating scalp and hair disorders. Through magnified scalp visualization, dermatologists can assess clinically relevant features such as follicular density, hair-diameter variation, and miniaturized hairs. Despite its diagnostic value, trichoscopic analysis remains largely dependent on manual interpretation, which introduces subjectivity and inter-observer variability.

The clinical logic of the proposed pipeline is measurement-driven. Follicular-unit detection determines where individual units are located; ordinal follicle-count classification estimates whether each unit contains one, two, or three-plus visible shafts; and calibrated width analysis characterizes shaft miniaturization. These measurements are then summarized by anatomical region. For the AGA use case, frontal measurements are interpreted relative to an occipital reference because patterned miniaturization, increased single-hair units, and increased vellus-hair burden are established trichoscopic indicators of AGA~\cite{ref1,ref7,ref9,ref10,ref11,ref12,ref13}. Thus, each computer-vision stage provides a specific quantitative input to the final clinical decision-support layer rather than acting as an isolated prediction task.

Quantitative analysis remains challenging because follicular units occupy only a small fraction of a full image and may be densely packed. Natural shaft overlap and crossing can obscure follicular openings and proximal shaft boundaries, while dandruff can introduce additional scalp texture around the regions of interest. Because image acquisition was standardized, the analysis focuses on these content-related challenges.

To address these constraints, the detector uses a slicing-aided strategy inspired by SAHI(Slicing Aided Hyper Inference), in which the scalp region is divided into overlapping tiles so that small follicular units occupy a larger fraction of the detector input~\cite{ref6}. A handcrafted support map is used in the count-classification and shaft-tracing stages to emphasize dark filamentous structures, while the root-anchored width method traces candidate shafts outward from a detected follicular-unit location instead of relying on unattributed global skeleton branches.

Recent advances in deep learning and computer vision have shown strong potential in medical image analysis~\cite{ref2,ref3}. YOLO-family detectors are attractive for small-object localization because they combine strong detection performance with single-stage inference, while EfficientNet backbones offer favorable accuracy-to-computation trade-offs for fine-grained classification. Classical image-processing methods remain valuable for thin filamentous structures when explicit geometric measurement is required.

The primary contribution of this work is therefore a systems-level integration of complementary techniques into an interpretable quantitative scalp-analysis framework. The framework is not presented as a stand-alone learned AGA-vs.-non-AGA classifier. Instead, it automates measurements that can support dermatologists in evaluating AGA and, with appropriately redefined clinical criteria, may be extended to other scalp and hair disorders.

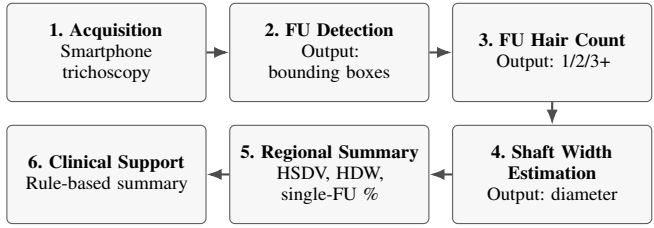
\begin{figure}[t]
\centering
\begin{tikzpicture}[
  node distance=3mm and 3mm,
  stage/.style={draw=black!65, rounded corners=2pt, align=center, minimum height=13mm, text width=24mm, fill=black!3, font=\scriptsize},
  arr/.style={-{Latex[length=2mm]}, thick, draw=black!70}
]
\node[stage] (a) {\textbf{1. Acquisition}\\Smartphone trichoscopy};
\node[stage, right=of a] (b) {\textbf{2. FU Detection}\\Output: bounding boxes};
\node[stage, right=of b] (c) {\textbf{3. FU Hair Count}\\Output: 1/2/3+};
\node[stage, below=of c] (d) {\textbf{4. Shaft Width Estimation}\\Output: diameter};
\node[stage, left=of d] (e) {\textbf{5. Regional Summary}\\HSDV, HDW, single-FU \%};
\node[stage, left=of e] (f) {\textbf{6. Clinical Support}\\Rule-based summary};
\draw[arr] (a) -- (b);
\draw[arr] (b) -- (c);
\draw[arr] (c) -- (d);
\draw[arr] (d) -- (e);
\draw[arr] (e) -- (f);
\end{tikzpicture}
\caption{Compact high-level workflow. Each stage passes interpretable measurements to the next module, ending with a clinician-facing rule-based summary.}
\label{fig:overview}
\end{figure}

\section{Related Work}
Computational approaches to scalp-image analysis have progressed from classical phototrichogram techniques~\cite{ref4} and handcrafted feature pipelines~\cite{ref5} to deep learning methods that learn visual representations directly from data~\cite{ref3}. Classical thresholding, edge detection, and morphological processing can perform well under controlled conditions, but dense shaft overlap, small target scale, and complex scalp texture remain challenging. Slicing-aided inference has been shown to improve small-object recall in high-resolution images~\cite{ref6}, and end-to-end AI workflows for microscopic hair imaging and scalp disorders have also been reviewed~\cite{ref2}.

The present work does not claim a novel detector or classifier backbone. Its contribution is the integration of follicular-unit localization, ordinal visible-shaft counting, calibrated shaft-width measurement, and multi-region clinical aggregation in one interpretable workflow. The current study emphasizes a working, computationally practical system and therefore benchmarks several YOLO and EfficientNet configurations internally rather than exhaustively comparing every detector, Vision Transformer, or CNN-transformer alternative. This is a scope limitation, and broader cross-family benchmarking is identified as future work.

\section{Dataset and Annotation}
\subsection{Clinical Cohort and Image Acquisition}
The associated clinical cohort comprised 243 patients with clinician-assigned diagnoses: 127 patients with AGA and 116 patients without AGA. Within the AGA group, 102 patients had female-pattern hair loss (FPHL) and 25 had male-pattern hair loss (MPHL). The non-AGA comparison group included 63 patients with telogen effluvium (TE) and 53 clinically normal patients. The AGA group contained 102 females and 25 males, whereas the non-AGA group contained 104 females and 12 males. Mean age was 29.15 years in the AGA group and 26.58 years in the non-AGA group, with median ages of 27 and 26 years, respectively. Written informed consent was obtained from participants, and the study was conducted under institutional ethics approval.

Clinical severity information was also available for the AGA group. Sinclair grading was recorded for 97 AGA patients (grade 1: 42; grade 2: 43; grade 3: 8; grade 4: 4). Norwood--Hamilton staging was recorded for 25 AGA patients (I: 4; II: 10; III: 5; III vertex: 1; IV: 1; V: 1; VI: 2; VII: 1). These clinician-provided labels and demographic summaries characterize the broader clinical cohort but are kept distinct from the module-level computer-vision evaluation described below.

\begin{table}[t]
\centering
\caption{Clinical cohort demographics and diagnosis composition.}
\label{tab:clinicalcohort}
\scriptsize
\begin{tabular}{lcc}
\toprule
Characteristic & AGA & Non-AGA\\
\midrule
Patients & 127 & 116\\
Female / male & 102 / 25 & 104 / 12\\
Mean age (years) & 29.15 & 26.58\\
Median age (years) & 27 & 26\\
FPHL & 102 & 0\\
MPHL & 25 & 0\\
Telogen effluvium & 0 & 63\\
Clinically normal & 0 & 53\\
\bottomrule
\end{tabular}
\end{table}

The computer-vision experiments use an expert-annotated imaging dataset of 160 patients. Images were captured in a low-resource acquisition setup using a handheld dermatoscope (DermaSight DS175) in cross-polarized mode to reduce surface glare, at $10\times$ magnification, with a 12MP iPhone 7 smartphone camera mounted using a magnetic connector. A specially designed spacer containing a $16\times 6.25$~mm field slit was used, and images were acquired at a native resolution of $3024\times4032$ pixels. The slit was centered along the axis of hair parting. For each patient in this imaging dataset, three photographs were obtained at five anatomical landmarks (frontal, vertex, left temporal, right temporal, and occipital), producing 15 images per patient and 2,400 trichoscopic frames overall.

Because each patient contributes multiple images, a strict patient-disjoint 70/15/15 split is used within the 160-patient expert-annotated imaging dataset. One hundred twelve patients (1,680 images) are assigned to training, 24 patients (360 images) to validation, and 24 patients (360 images) to testing; all images and all derived follicular-unit crops from one patient remain in the same partition.

All images in the expert-annotated imaging dataset were acquired using a standardized protocol with the same dermatoscope, smartphone camera, magnification, spacer geometry, and anatomical acquisition procedure. The main naturally occurring visual obstacle identified by the clinical team is dandruff, which is a clinical scalp finding and is well represented in the dataset. Geometric and photometric augmentation is used during training as regularization rather than as compensation for inconsistent acquisition conditions.

The image and annotation statistics used for reproducibility are summarized in Table~\ref{tab:datasetstats}. De-identified clinical summaries and an annotated image subset may be made available upon reasonable request, subject to institutional ethics, consent, and data-sharing approval.

\begin{table}[t]
\centering
\caption{Computer-vision dataset composition and derived task subsets.}
\label{tab:datasetstats}
\small
\begin{tabular}{lc}
\toprule
Statistic & Count\\
\midrule
Expert-annotated imaging patients & 160\\
Images per imaging patient & 15\\
Total trichoscopic images & 2,400\\
Anatomical landmarks & 5\\
Images per landmark & 480\\
Expert FU annotations & $\approx$158,000\\
Classifier-ready FU crops & 126,626\\
Additional end-to-end clinical images & 500\\
\bottomrule
\end{tabular}
\end{table}

Per-region FU annotation counts require an additional clinical audit before they can be treated as finalized ground-truth statistics. They are therefore not used as quantitative evidence in this study; producing these expanded image-level summaries is part of the planned dataset-release preparation.

\subsection{Annotation and Derived Task Datasets}
Each visible follicular unit (FU) was manually delineated by an expert trichologist using an axis-aligned bounding box and assigned one of three ordinal visible shaft-count labels: one, two, or three-plus. Across the dataset, approximately 158,000 FU instances were annotated.

\begin{figure}[t]
\centering
\includegraphics[width=0.72\columnwidth]{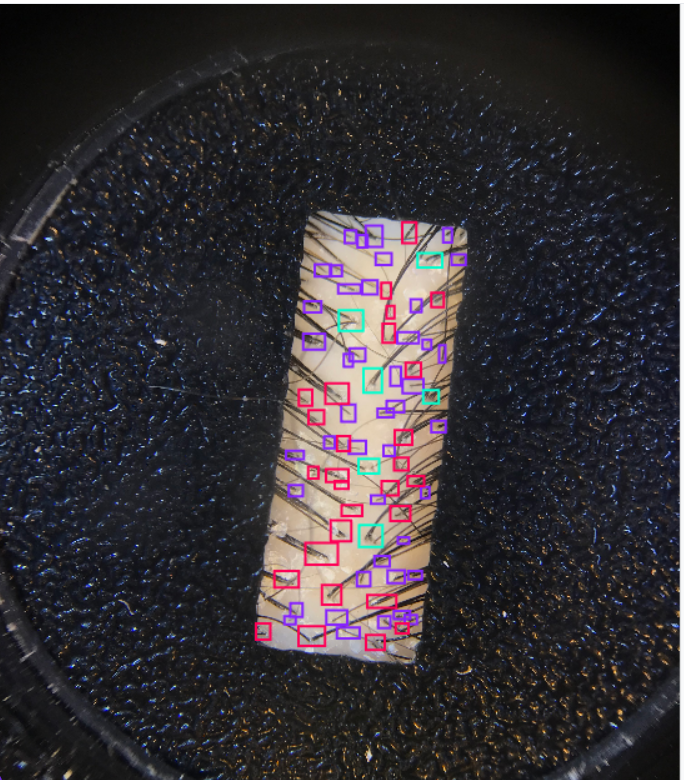}
\caption{Representative trichoscopic image with expert follicular-unit bounding boxes.}
\label{fig:annotated}
\end{figure}

For the follicle-count classification task, each expert bounding box was expanded by 10 pixels on all sides, and crops smaller than 16 pixels in either dimension were discarded. The original filtering procedure yielded 126,626 classifier-ready crops. Crop assignment follows the parent patient's split rather than an independent crop- or image-level split.

\begin{figure}[t]
\centering
\begin{tabular}{ccc}
\includegraphics[width=0.27\columnwidth]{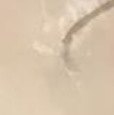} &
\includegraphics[width=0.27\columnwidth]{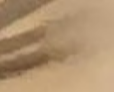} &
\includegraphics[width=0.27\columnwidth]{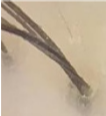}\\[-1mm]
\scriptsize One hair & \scriptsize Two hairs & \scriptsize Three-plus
\end{tabular}
\caption{Representative expert-annotated follicular-unit crops for the ordinal count task.}
\label{fig:crops}
\end{figure}

A separate set of 500 additional clinical trichoscopic images supplied by a collaborating dermatologist was reserved for system-level end-to-end processing. In this set, YOLO-generated boxes are propagated directly to the classifier and morphometric modules. Because the set does not contain independent dermatologist-verified one/two/three-plus labels for every detector-generated crop, it is not used to compute a second crop-level classification accuracy; instead, it is used to examine the operation of the fully automated chain and the resulting regional summaries.

\section{Methodology}
\subsection{Overall Pipeline}
The system contains four quantitative stages followed by a clinical aggregation stage: (1) FU localization, (2) ordinal visible-shaft counting, (3) hair-shaft width estimation and hair-type categorization, and (4) regional feature aggregation. The final rule-based layer maps those features to an interpretable AGA-support summary. Figure~\ref{fig:overview} shows the information passed between stages.

\subsection{Follicular-Unit Detection}
\subsubsection{Task objective and input}
FU localization is formulated as a single-class small-object detection problem. The goal is to identify candidate follicular units with sufficient spatial coverage for downstream count and width analysis.

\subsubsection{Preprocessing}
Each image first undergoes a lightweight region-of-interest crop to isolate the scalp-bearing field. The grayscale image is thresholded, the largest tissue-bearing contour is extracted, and its bounding rectangle is expanded by a 20-pixel margin. Degenerate ROIs revert to the original image.

After ROI extraction, the image is divided into overlapping $640\times640$ tiles using a 30\% overlap. This SAHI-style strategy makes small FUs occupy a larger fraction of the detector input and preserves follicles near tile boundaries. CLAHE is applied to luminance to improve local contrast, and the number of background-only tiles retained per source image is capped.

\begin{figure}[t]
\centering
\includegraphics[width=0.42\columnwidth]{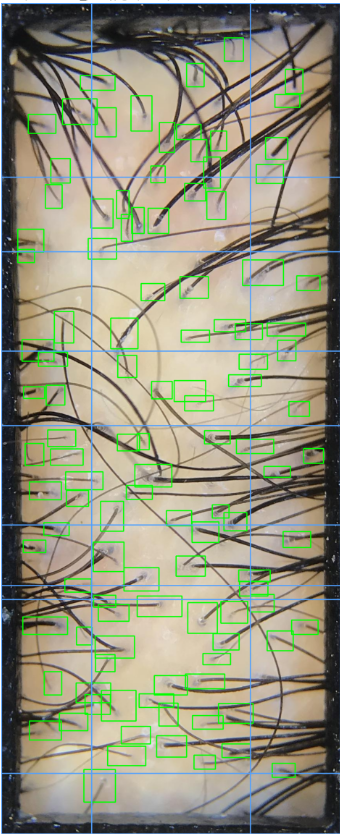}
\caption{Slicing-aided preprocessing: overlapping tiles increase the relative scale of small follicular units.}
\label{fig:sahi}
\end{figure}

\subsubsection{Model choice and architecture}
YOLO was selected because its single-stage architecture offers a practical combination of small-object localization quality and inference speed for clinical deployment. Model selection experiments identified the medium-scale YOLOv8 model (YOLOv8m) as the final detector, providing strong patient-level localization performance with moderate model complexity. All FU subtypes are collapsed into one detection class because count subtype is handled by the subsequent ordinal classifier.

\subsubsection{Output and implementation}
Detector outputs consist of FU bounding boxes and confidence scores. During training, AdamW optimization, image size 640, and geometric/photometric augmentation are used, including horizontal/vertical flips, rotations, scaling, HSV jitter, mosaic, mixup, and copy-paste. The operating confidence threshold is selected on the validation set and then fixed for held-out test evaluation.

\subsection{Ordinal Follicular-Count Classification}
\subsubsection{Task objective and evaluation distinction}
Each FU is classified into one of three ordered categories: one, two, or three-plus visible hairs. Two uses of bounding boxes must be distinguished. For controlled module-level training and evaluation, crops are generated from expert-annotated boxes because these are the only boxes with dermatologist/trichologist-verified one/two/three-plus ground-truth labels. In actual automated operation, crops are generated from YOLO predictions. The reported 87\% classifier accuracy therefore measures count classification under ground-truth localization and should not be interpreted as end-to-end system accuracy.

\subsubsection{Support-map generation}
The RGB crop is supplemented by a single-channel support map designed to emphasize thin dark shaft-like structures. Let $Y$ be normalized luminance after contrast normalization. We construct five normalized cues: global darkness $D_g=1-Y$; local darkness contrast $D_l=\max(0,\mathrm{blur}(Y)-Y)$, where the blurred image estimates the local scalp background; Sobel edge magnitude $E$ computed from the luminance gradient; morphological black-hat response $B$ using line- and ellipse-shaped structuring elements matched to thin shafts; and a second-order ridge response $R$ computed on inverted luminance at multiple small scales. Each cue is min--max normalized inside the crop after percentile clipping to reduce the influence of isolated extreme responses.

The support map is
\begin{equation}
S=\mathcal{N}\left(w_gD_g+w_lD_l+w_eE+w_bB+w_rR\right),
\end{equation}
where $w_i\geq 0$ are fixed cue weights and $\mathcal{N}(\cdot)$ rescales the result to $[0,1]$. The same fixed support-map generator is used for both classification and width estimation so that the auxiliary channel consistently highlights dark, elongated, locally contrasted structures. The support map is not a replacement for the RGB appearance; it is an auxiliary structural prior that increases the salience of filamentous shaft trajectories while leaving color and texture information available to the RGB branch.

\subsubsection{Architecture and model choice}
EfficientNet-B2, B3, B5, and support-map augmented variants were explored. EfficientNet was prioritized because of its accuracy-per-parameter efficiency, suitability for fine-grained image classification, and compatibility with the ordinal CORAL head~\cite{ref8}. Transformer and hybrid backbones were not exhaustively benchmarked because the study prioritized an integrated working system under limited compute; broader backbone comparisons are left for future work.

The final EfficientNet-B5+S configuration resizes RGB and support-map inputs to $456\times456$ and concatenates them into a four-channel tensor. An ImageNet-pretrained EfficientNet-B5 backbone is followed by dropout ($p=0.3$) and a CORAL projection to $K-1$ rank logits for $K=3$ ordered classes. Inverse-frequency weighted sampling mitigates class imbalance. AdamW with differential backbone/head learning rates, mixed precision, gradient clipping, and OneCycle scheduling is used.

\subsubsection{Output passed downstream}
The classifier outputs an ordinal label corresponding to one, two, or three-plus visible shafts for each FU crop. This predicted count is passed to the width-estimation stage as a soft structural constraint: it determines how many proximal shaft candidates should be retained when several plausible paths emerge from the same follicular unit.

\subsection{Hair-Width Estimation and Hair-Type Classification}
\subsubsection{Calibration and task objective}
The valid imaging field is detected from a grayscale, Gaussian-smoothed, Otsu-thresholded mask followed by morphological closing and a rotated-rectangle fit. The ROI is perspective-warped into an upright crop, and detector boxes are transformed into the same coordinate system. Pixel distances are converted to micrometers using the known $16\times6.25$~mm dimensions of the spacer slit.

\subsubsection{Global branch-based baseline}
The baseline generates a global hair mask using CLAHE-enhanced luminance, multi-scale black-hat filtering, Frangi ridge response, Otsu thresholding, and morphological cleanup. Local FU crops are reprocessed with the same hair-enhancement filters and inserted back into the corresponding global coordinates by taking the pixelwise maximum of the global and local mask probabilities before final thresholding. This fusion step is intended to recover short or low-contrast strands that are suppressed when the full image is processed at once.

The fused mask is skeletonized to a one-pixel-wide centerline and converted into connected branches. For each detected FU, candidate branches are collected from a search window around the FU center. Branches are ranked by three criteria: distance between the branch endpoint and the FU root, outward extension away from the box center, and local mask/support strength along the proximal branch. Branches that pass through the FU without a nearby endpoint or that primarily overlap another FU search region are discarded. Width is estimated from proximal shaft geometry by sampling intensity or mask profiles perpendicular to the selected centerline.

\subsubsection{Root-anchored support-map method}
The proposed method treats each detected FU box as an anatomical root anchor. For every FU, a local analysis crop is extracted around the box with contextual padding, transformed into calibrated coordinates, and converted to a support map using the same normalized darkness, contrast, black-hat, edge, and ridge cues described above. Several local threshold candidates are evaluated, including an adaptive threshold derived from the crop histogram and stricter percentile-based thresholds. A candidate threshold is accepted when it produces connected shaft components that intersect the root/core region while preserving outward branches beyond the FU box; otherwise the next threshold candidate is tested.

The selected binary mask is skeletonized into a graph whose nodes correspond to endpoints, junctions, and sampled centerline pixels. Candidate paths are seeded from skeleton pixels inside or immediately adjacent to the FU root region and are traced outward along the graph. Paths are rejected if they enter the crop from one side and pass through the FU without terminating near the root, if they duplicate a higher-scoring path from the same root, if they bend sharply over a short distance, or if their support-map response is unstable along the proximal segment. Remaining paths are scored to favor smooth continuation, increasing distance from the root center, and high support-map response while penalizing pass-through, duplicate, and unstable trajectories. Conceptually,
\begin{equation}
Q(p)=\lambda_s q_{\mathrm{smooth}}+\lambda_o q_{\mathrm{outward}}+\lambda_r q_{\mathrm{support}}-\lambda_p q_{\mathrm{penalty}},
\end{equation}
with fixed non-negative weights. The expected number of visible strands is guided by the ordinal FU count.

Width is measured perpendicular to the selected centerline near the proximal shaft, primarily around 250~$\mu$m from the root. A short segment around this anchor point is sampled at multiple neighboring centerline locations, and the final diameter is taken only when the perpendicular profiles have consistent edge-to-edge distances and sufficient local support-map contrast. Local profile consistency checks reject unstable measurements caused by crossing hairs, local shaft overlap, or dandruff near the measurement path. Diameters are categorized as vellus ($<30~\mu$m), intermediate (30--60~$\mu$m), or terminal ($>60~\mu$m)~\cite{ref9}. The per-shaft diameters and type labels are passed to the regional aggregation stage.

\subsection{Multi-Region Clinical Decision Support}
The final layer is deliberately rule-based rather than a learned AGA-vs.-non-AGA classifier. Its role is to organize automatically measured trichoscopic features into an interpretable summary for dermatologist review. For AGA, frontal measurements are compared with the occipital reference because AGA is a patterned miniaturization disorder~\cite{ref1,ref7,ref10,ref11}.

For each region, the system computes the percentage of single-hair FUs, hair distribution width (HDW), and hair-shaft diameter variation (HSDV). Let $D_i$ be the diameter of the $i$th measured shaft, $n$ the number of measured shafts, and $n_v$ the number classified as vellus. Then
\begin{align}
\bar{D}&=\frac{1}{n}\sum_{i=1}^{n}D_i,\\
\sigma_D&=\sqrt{\frac{\sum_{i=1}^{n}(D_i-\bar{D})^2}{n-1}},\\
\mathrm{HDW}&=\frac{\sigma_D}{\bar{D}}\times100,\\
\mathrm{HSDV}&=\frac{n_v}{n}\times100.
\end{align}
Following the clinical literature, the AGA use case evaluates
\begin{align}
C_1 &: \mathrm{HSDV}_{\mathrm{frontal}}>20\%,\\
C_2 &: P_{1,\mathrm{frontal}}>2P_{1,\mathrm{occipital}},
\end{align}
where $P_1$ is the percentage of single-hair FUs. Increased vellus burden and frontal predominance of single-hair units are clinically meaningful because they reflect miniaturization and reduced multi-hair follicular-unit density~\cite{ref1,ref9,ref11,ref12,ref13}.

\begin{table}[t]
\centering
\caption{Clinician-facing AGA support categories.}
\label{tab:rules}
\small
\begin{tabular}{lll}
\toprule
Output & Rule & Interpretation\\
\midrule
Consistent with AGA & $C_1\land C_2$ & Both criteria support AGA\\
Possible AGA & $C_1\oplus C_2$ & One criterion positive\\
AGA not supported & $\neg C_1\land\neg C_2$ & Neither criterion positive\\
\bottomrule
\end{tabular}
\end{table}

These outputs are not intended as autonomous diagnoses. The system also displays regional counts, density, HDW, HSDV, and shaft-width distributions so that the dermatologist can stage or grade the observed scalp changes in the full clinical context.

\section{Experimental Results}
\subsection{Follicular-Unit Detection}
Table~\ref{tab:yolo_comparison} summarizes the initial YOLO-family screening experiments performed with smart-crop and SAHI-style tiled preprocessing. These experiments were used to identify a strong baseline and motivated further focused development within the YOLOv8 family. After additional annotation cleaning, data inspection, and optimization under the patient-disjoint protocol, the medium-scale YOLOv8 model (YOLOv8m) achieved a held-out test mAP@0.5 of 0.920 with recall of 0.860 and was selected for the final pipeline. Thus, the table documents the earlier model-screening stage, whereas the reported YOLOv8m result reflects the subsequent focused YOLOv8 optimization.

\begin{table}[t]
\centering
\caption{Initial YOLO-family follicular-unit screening results.}
\label{tab:yolo_comparison}
\scriptsize
\setlength{\tabcolsep}{2.0pt}
\resizebox{\columnwidth}{!}{%
\begin{tabular}{llcccccc}
\toprule
Model & Split & P & R & F1 & mAP@0.5 & mAP@0.5:0.95 & Params \\
\midrule
\multirow{2}{*}{YOLOv8l-P2} & Val  & 0.672 & 0.712 & 0.691 & 0.684 & 0.234 & 42.8M \\
 & Test & 0.681 & 0.707 & 0.694 & 0.694 & 0.233 & -- \\
\multirow{2}{*}{YOLO11s-P2} & Val  & 0.673 & 0.687 & 0.680 & 0.565 & 0.166 & 12.5M \\
 & Test & 0.686 & 0.699 & 0.692 & 0.587 & 0.173 & -- \\
\multirow{2}{*}{YOLO26s-P2} & Val  & 0.560 & 0.577 & 0.568 & 0.486 & 0.143 & 9.4M \\
 & Test & 0.570 & 0.580 & 0.575 & 0.495 & 0.143 & -- \\
\multirow{2}{*}{YOLO26m-P2} & Val  & 0.585 & 0.584 & 0.585 & 0.500 & 0.146 & 20.3M \\
 & Test & 0.583 & 0.594 & 0.588 & 0.512 & 0.149 & -- \\
\multirow{2}{*}{YOLO26l-P2} & Val  & 0.637 & 0.620 & 0.628 & 0.563 & 0.172 & 25.0M \\
 & Test & 0.634 & 0.631 & 0.633 & 0.570 & 0.169 & -- \\
\bottomrule
\end{tabular}%
}
\end{table}

\subsection{Follicular-Count Classification}
Six EfficientNet variants were evaluated for the three-class ordinal task. The support-map augmented EfficientNet-B5 achieved the highest controlled crop-level performance: 87.0\% accuracy, macro precision 0.86, macro recall 0.85, and macro F1 0.85.

\begin{table}[t]
\centering
\caption{EfficientNet classifier comparison under expert-verified FU localization. ``+S'' denotes inclusion of the support map as a fourth input channel.}
\label{tab:eff_comparison}
\scriptsize
\setlength{\tabcolsep}{3.2pt}
\begin{tabular}{lccccc}
\toprule
Model & Acc. & Macro P & Macro R & Macro F1 & Val. Acc.\\
\midrule
EfficientNet-B2   & 0.827 & 0.81 & 0.82 & 0.82 & 0.819\\
EfficientNet-B2+S & 0.835 & 0.83 & 0.82 & 0.82 & 0.832\\
EfficientNet-B3   & 0.839 & 0.83 & 0.83 & 0.83 & 0.832\\
EfficientNet-B3+S & 0.842 & 0.83 & 0.83 & 0.83 & 0.831\\
EfficientNet-B5   & 0.860 & 0.84 & 0.85 & 0.85 & 0.845\\
\textbf{EfficientNet-B5+S} & \textbf{0.870} & \textbf{0.86} & \textbf{0.85} & \textbf{0.85} & \textbf{0.849}\\
\bottomrule
\end{tabular}
\end{table}

For EfficientNet-B5+S, class-wise F1 scores were 0.92, 0.84, and 0.81 for one-, two-, and three-plus-hair units, respectively.

Across the evaluated backbones, adding the support-map channel produced small but consistent accuracy gains (B2: 0.827 to 0.835; B3: 0.839 to 0.842; B5: 0.860 to 0.870), with EfficientNet-B5+S providing the strongest overall result.

These values quantify the intrinsic visible-shaft counting capability when a true FU has already been localized by an expert. They do not include detector false positives, detector misses, or localization offsets. This distinction is explicitly separated because the available expert annotations provide count labels for manually defined FUs, not for every YOLO-generated proposal.

\subsection{End-to-End Processing on the Additional 500-Image Set}
The additional 500 clinical images were processed using the complete automated sequence: YOLOv8m detection, predicted-box cropping, EfficientNet-B5+S count prediction, root-anchored width analysis, regional aggregation, and the clinical support rules. This experiment evaluates system-level operation with real detector outputs rather than manually supplied boxes. However, a detector-generated crop cannot be assigned a valid count-classification accuracy unless an expert independently determines whether the proposal is a true FU and, if so, assigns its one/two/three-plus label. Because such per-proposal expert labels are not available for all 500 images, we do not report an artificial end-to-end crop accuracy. In comparison with human-expert annotations, the automated pipeline achieved mean absolute errors (MAE) of 6.56 for the follicle-detection task and 16.59 for the follicle-classification task on this 500-image set. These MAE results characterize system-level agreement at the task-output level and are reported separately from the controlled expert-box classifier benchmark.

\subsection{Hair-Width Estimation}
The width study compared a global branch-based method with the root-anchored support-map method using a preliminary single-expert reference. The root-anchored method produced 158 valid measurements from 160 reference strands and had an average runtime of 12.0~s/image. The global branch-based method was much slower (142.8~s/image) and its agreement analysis was completed on only a 15-strand subset.

\begin{table}[t]
\centering
\caption{Exploratory width-estimation results from the manual-reference experiment.}
\label{tab:width}
\scriptsize
\begin{tabular}{lcccc}
\toprule
Method & Valid & ICC & Bias ($\mu$m) & Time (s/img)\\
\midrule
Global branch-based & 15/15 & 0.818 & +0.8 & 142.8\\
Root-anchored support-map & 158/160 & 0.905 & +4.087 & 12.0\\
\bottomrule
\end{tabular}
\end{table}

Because the two agreement statistics were computed on unequal subsets, they are not presented as a controlled head-to-head accuracy comparison. The main justified conclusion from this experiment is the large runtime advantage of the root-anchored implementation and its feasibility on a much larger reference set.

\subsection{Clinical Output and Visualization}
\begin{figure}[t]
\centering
\includegraphics[width=0.98\columnwidth]{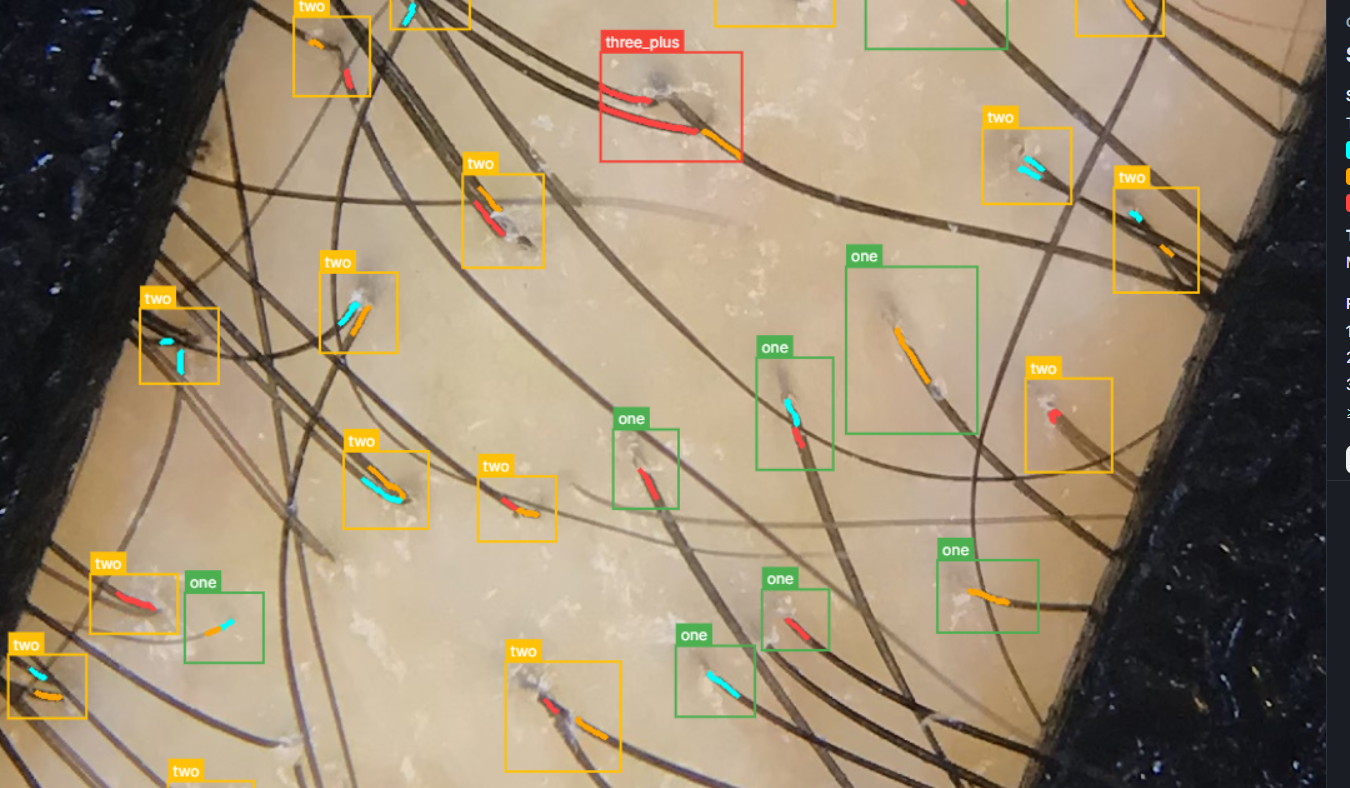}
\caption{Representative end-to-end output. FU boxes are color-coded by predicted ordinal count and measured shaft segments are rendered according to diameter category.}
\label{fig:endtoend}
\end{figure}

The rule layer summarizes regional measurements and frontal-to-occipital comparisons into one of the support categories in Table~\ref{tab:rules}; final diagnosis and formal staging remain the dermatologist's responsibility. Clinician-assigned labels are available for the broader 243-patient cohort (127 AGA, 116 non-AGA), but automated rule outputs have not yet been paired with all patient-level diagnoses; therefore diagnostic accuracy, sensitivity, specificity, and kappa are not reported.

\section{Discussion}
Patient-disjoint evaluation reduces the risk of identity leakage. Initial YOLO-family screening was followed by focused YOLOv8 development, with YOLOv8m reaching mAP@0.5=0.920 and recall=0.860. EfficientNet-B5+S achieved 87\% expert-box count accuracy (macro F1=0.85), while the 500-image predicted-box experiment yielded system-level MAE of 6.56 for detection and 16.59 for classification.

The measured features are clinically interpretable: increased one-hair FU proportion reflects loss of multi-hair density, while vellus burden and diameter diversity capture miniaturization; frontal-to-occipital comparison reflects the patterned distribution of AGA. Acquisition was standardized, and the main content-related challenges are small/dense FUs, natural shaft overlap/crossing, and dandruff, which is represented in the dataset.

Limitations include preliminary width validation, unequal reference subsets for the two width methods, the small FotoFinder comparison, limited cross-family architecture benchmarking, and the absence of a completed patient-level comparison between rule outputs and all 243 clinician diagnoses. The framework should therefore be interpreted as quantitative clinical decision support rather than an autonomous disease classifier.

\section{Conclusion and Future Work}
We presented an interpretable multi-stage framework combining FU detection, ordinal shaft counting, calibrated root-anchored width measurement, and regional clinical aggregation. YOLOv8m achieved mAP@0.5=92\% and recall=86\%, while EfficientNet-B5+S achieved 87\% expert-box count accuracy (macro F1=0.85). Future work will prioritize larger FotoFinder/expert width validation, dermatologist-reviewed detector proposals and clinical outputs, broader backbone comparisons, multi-center evaluation, and condition-specific extension to other scalp disorders. De-identified data may be shared upon reasonable request subject to institutional approval.


\begin{thebibliography}{13}
\bibitem{ref1} A. Kuczara, A. Waskiel-Burnat, A. Rakowska, M. Olszewska, and L. Rudnicka, ``Trichoscopy of androgenetic alopecia: A systematic review,'' \emph{J. Clin. Med.}, vol. 13, no. 7, p. 1962, Mar. 2024, doi: 10.3390/jcm13071962.
\bibitem{ref2} C. Gong, Y. Su, S. Ruan, and H. Hu, ``Artificial intelligence in microscopic hair imaging for scalp disorders: From image acquisition to clinical decisions,'' \emph{Med. Image Anal.}, vol. 111, p. 104017, Jun. 2026, doi: 10.1016/j.media.2026.104017.
\bibitem{ref3} A. Esteva et al., ``A guide to deep learning in healthcare,'' \emph{Nat. Med.}, vol. 25, no. 1, pp. 24--29, Jan. 2019, doi: 10.1038/s41591-018-0316-z.
\bibitem{ref4} R. Hoffmann, ``TrichoScan: A novel tool for the analysis of hair growth in vivo,'' \emph{J. Investig. Dermatol. Symp. Proc.}, vol. 8, no. 1, pp. 109--115, Jun. 2003, doi: 10.1046/j.1523-1747.2003.12183.x.
\bibitem{ref5} N. Debnath, N. Das, S. Sarkar, and M. Nasipuri, ``An automated approach towards digital photo-trichogram for hair fall diagnosis,'' in \emph{Proc. 52nd Annu. Conv. Comput. Soc. India (CSI 2017)}, 2018, pp. 314--329.
\bibitem{ref6} F. C. Akyon, S. O. Altinuc, and A. Temizel, ``Slicing aided hyper inference and fine-tuning for small object detection,'' in \emph{Proc. IEEE Int. Conf. Image Process. (ICIP)}, 2022, pp. 966--970, doi: 10.1109/ICIP46576.2022.9897990.
\bibitem{ref7} L. Rudnicka, A. Rakowska, and M. Olszewska, ``Trichoscopy: How it may help the clinician,'' \emph{Dermatol. Clin.}, vol. 31, no. 1, pp. 29--41, Jan. 2013.
\bibitem{ref8} W. Cao, V. Mirjalili, and S. Raschka, ``Rank consistent ordinal regression for neural networks with application to age estimation,'' \emph{Pattern Recognit. Lett.}, vol. 140, pp. 325--331, Dec. 2020.
\bibitem{ref9} O. de Lacharri\`ere et al., ``Hair diameter diversity: A clinical sign reflecting the follicle miniaturization,'' \emph{Arch. Dermatol.}, vol. 137, no. 5, pp. 641--646, May 2001.
\bibitem{ref10} A. Rakowska, M. Slowinska, E. Kowalska-Oledzka, M. Olszewska, and L. Rudnicka, ``Dermoscopy in female androgenic alopecia: Method standardization and diagnostic criteria,'' \emph{Int. J. Trichology}, vol. 1, no. 2, pp. 123--130, Jul. 2009.
\bibitem{ref11} S. Inui, ``Trichoscopy for common hair loss diseases: Algorithmic method for diagnosis,'' \emph{J. Dermatol.}, vol. 38, no. 1, pp. 71--75, Jan. 2011.
\bibitem{ref12} A. Majd, L. D. N. Santos, T. W. Chu, J. Shapiro, H. Lui, and T. K. Lee, ``Hair distribution width--a novel trichoscopy parameter for hair miniaturization in androgenetic alopecia,'' \emph{J. Eur. Acad. Dermatol. Venereol.}, vol. 33, no. 10, pp. e371--e372, Oct. 2019.
\bibitem{ref13} H. U. Park, K. B. Chung, and D.-Y. Kim, ``Quantitative measurement of hair diameter diversity as a diagnostic indicator of androgenetic alopecia in Korean males: A cross-sectional study,'' \emph{JAAD Int.}, vol. 15, pp. 121--126, Jun. 2024.
\end{thebibliography}
\end{document}